\documentclass[format=acmsmall, review=false, screen=true]{acmart}
\usepackage{booktabs} 
\usepackage[normalem]{ulem}

\copyrightyear{2018}

\setcopyright{acmlicensed}
\acmYear{2018}
\acmJournal{PACMHCI}

\newcommand{\jacob}[1]{\textcolor{blue}{#1 -JE}}

\usepackage{comment}

\usepackage[english]{babel}
\usepackage[utf8]{inputenc}
\usepackage{algorithm}
\usepackage[noend]{algpseudocode}

\usepackage{multirow}
\usepackage{enumitem}
\usepackage{booktabs}
\usepackage{dcolumn} 
\usepackage{array} 
\usepackage{cleveref}
\crefname{section}{§}{§§}
\Crefname{section}{§}{§§}
\usepackage{courier}
\usepackage{soul,color}
\definecolor{lyellow}{rgb}{0.98, 0.91, 0.71}
\sethlcolor{lyellow}

\begin{document}
\title[Writing Style Norm Convergence in Wikipedia]{Mind Your POV: Convergence of Articles and Editors Towards Wikipedia's Neutrality Norm}

\author{Umashanthi Pavalanathan}
\affiliation{%
  \institution{Georgia Institute of Technology}
    \department{School of Interactive Computing}
   \country{USA}
}

\author{Xiaochuang Han}
\affiliation{%
  \institution{Georgia Institute of Technology}
    \department{College of Computing}
   \country{USA}
}

\author{Jacob Eisenstein}
\affiliation{%
  \institution{Georgia Institute of Technology}
  \department{School of Interactive Computing}
   \country{USA}
}

\begin{abstract}
Wikipedia has a strong norm of writing in a ``neutral point of view'' (NPOV). Articles that violate this norm are tagged, and editors are encouraged to make corrections. But the impact of this tagging system has not been quantitatively measured. Does NPOV tagging help articles to converge to the desired style? Do NPOV corrections encourage editors to adopt this style? We study these questions using a corpus of NPOV-tagged articles and a set of lexicons associated with biased language. An interrupted time series analysis shows that after an article is tagged for NPOV, there is a significant decrease in biased language in the article, as measured by several lexicons. However, for individual editors, NPOV corrections and talk page discussions yield no significant change in the usage of words in most of these lexicons, including Wikipedia's own list of ``words to watch.'' This suggests that NPOV tagging and discussion does improve content, but has less success enculturating editors to the site's linguistic norms.
\end{abstract}

%
%

\ccsdesc[300]{Human-centered computing~Empirical studies in collaborative and social computing}

%
%

\keywords{Wikipedia, online communities, community norms, norm enforcement, objective language, writing style.}

\maketitle

\renewcommand{\shortauthors}{U. Pavalanathan et al.}

\section{Introduction}
Wikipedia is the largest online collaborative content creation community. The English Wikipedia contains nearly 5.6 million articles and nearly 140 thousand actively contributing accounts~\cite{wiki2018wiki}. Similar to the norms or rules of other online communities~\cite{burnett2003beyond}, Wikipedia also has a set of norms regarding content and member participation~\cite{wiki2018norms}. The three core content policies of Wikipedia are: ``neutral point of view (NPOV)'', ``verifiability'', and ``no original research''~\cite{wiki2018npov}. Wikipedia provides strict guidelines about NPOV, and non-complying articles are marked with an NPOV tag by editors. Such NPOV-tagged articles are listed under the ``NPOV disputed'' Wikipedia category. While the NPOV tagging system has been in place for over a decade, its impact on Wikipedia has not been quantitatively measured. In this paper, we study the effectiveness of NPOV tagging as a language norm enforcement strategy in Wikipedia, measuring its impact on articles as well as individual editors. Specifically, we ask whether tagging helps articles and editors converge towards the community's prescribed writing style, as a step toward the larger goal of understanding how collaborative communities converge on language standards. 

\subsection{Norm Enforcement in Wikipedia} \label{subsec:wiki-norms}
Wikipedia provides detailed guidelines of the preferred writing styles through a manual of style~\cite{wiki2018style}. The manual of style contains guidelines not only for formatting such as capitalization, punctuations, and abbreviations, but also for grammar and vocabulary. The \emph{Neutral Point of View (NPOV)} guideline
states that ``\textit{All encyclopedic content on Wikipedia must be written from a neutral point of view (NPOV), which means representing fairly, proportionately, and, as far as possible, without editorial bias, all of the significant views that have been published by reliable sources on a topic.}''
The guideline further states that ``\textit{This policy is non-negotiable, and the principles upon which it is based cannot be superseded by other policies or guidelines, nor by editor consensus.}''  These editorial policies are enforced by the community of editors to maintain high quality of the articles. 

\begin{figure}[h]

\centering
\includegraphics[width=0.95\textwidth]{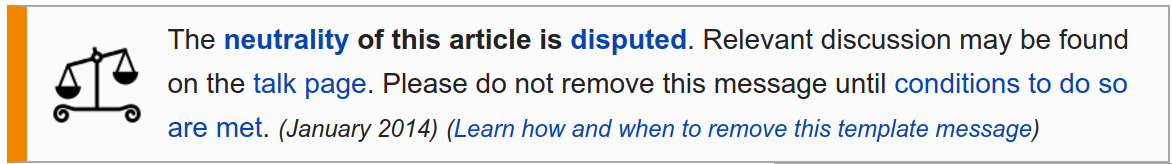}
\caption[Caption for Tag]{An NPOV tag displayed at the top of an article.\footnotemark}
\label{fig:npov-tag}
\end{figure}

\footnotetext{Image obtained from \url{https://en.wikipedia.org/wiki/Natural_Gas_Choice} on 30-August-2018.}
Wikipedia articles which do not follow the NPOV guidelines are marked by editors with tags such as \{\{POV\}\}, \{\{NPOV\}\}, and \{\{POV-section\}\} so that they can be included in the editing workflow (\autoref{fig:npov-tag}). From the explicit tagging of the non-NPOV content in Wikipedia articles and the availability of the content before and after NPOV revisions, we can develop a better understanding of the acceptable writing style in Wikipedia. The example below from our dataset shows a portion of a sentence before and after it was corrected for NPOV. Usage of \textbf{dictated} in (a) suggests a point of view and it is replaced with \textbf{overseen} in (b) to make the language more neutral. 
\begin{itemize}
\item[(a)] ``BL Republic is \textbf{dictated} by UberConsul Lars and High Marshall Bill ...''
\item[(b)] ``BL Republic is \textbf{overseen} by UberConsul Lars and High Marshall Bill ...''
\end{itemize}


\subsection{Research Questions and Findings}
We study the effects of NPOV tagging and revision as norm enforcement strategies to make Wikipedia an objective source of knowledge at the article level and at the editor level. 
\begin{itemize}
 \item[]\indent \textbf{{RQ1:} } \textbf{What are the article-level effects of NPOV tagging?}
 \item[]\indent \textbf{{RQ2:} } \textbf{What are the editor-level effects of NPOV correction?}
 \item[]\indent \textbf{\emph{RQ2a:} } {What effect does NPOV correction have on the writing style of the editors?}
 \item[]\indent \textbf{\emph{RQ2b:} } {What effect does NPOV correction have on the engagement of editors?}

\end{itemize}

We address these questions using a corpus of nearly 7,500 articles in the \textit{NPOV dispute} category, and we identify revisions in which a NPOV tag was added or removed, or in which content was corrected for NPOV. We then identify the first revision in which an article received a NPOV tag as the treatment point for the article. Similarly we identify editors whose content contributions were corrected for NPOV as treatment editors and identify the treatment revisions. As a measure of language, we use a set of lexicons associated with non-objective or biased language.\footnote{Lexical analysis is chosen because of the existence of well-validated lexicons, as described later. We leave for future work the analysis of bias in dimensions such as syntax, semantics, and pragmatics.} Considering editor contribution and article revisions before and after the treatment points, we analyze the effect of treatment using an interrupted time series analysis~\cite{bernal2017interrupted}. 

We find that the NPOV tagging helps articles to converge to the desired writing style of Wikipedia. However, for individual editors, NPOV corrections yield no statistically significant change in most lexicons used to characterize biased language, including Wikipedia's own ``words to watch'' lexicon. This finding holds even when NPOV corrections are paired with one-to-one interactions in the form of talk page discussions, which trigger email prompts to the corrected editor.
To summarize our results: NPOV norm enforcement helps to make the articles better, but does not lead individual editors to converge to the writing style norms of Wikipedia.

\subsection{Terminology}
Below, we provide a brief summary of different terminology used in this paper:
\begin{itemize}
\item[] \textbf{Edit} refers to any changes made to a Wikipedia article.
\item[] \textbf{Revision} is a version of a Wikipedia article resulted from editing the previous revision.
\item[] \textbf{Article} contains chronologically ordered revisions.
\item[] \textbf{Reverts} in Wikipedia refers to edits that restored the current revision to a previous revision. 
\item[] \textbf{NPOV tagging} refers to the addition of any NPOV templates to the articles. 
\item[] \textbf{NPOV correction} refers to a set of edits that are made to correct for NPOV errors. These are detected from the comment metadata in a revision.
\item[] \textbf{Treatment} refers to an intervention made to the subject of interest (i.e., article, editor).
\item[] \textbf{Pre-treatment} refers to the period prior to the treatment.
\item[] \textbf{Post-treatment} refers to the period after the treatment.
\item[] \textbf{Treatment to the articles} refers to the addition of an NPOV tag.
\item[] \textbf{Treatment to the editors} refers to editors' contribution being corrected for NPOV.
\item[] \textbf{Biased language} refers to words/phrases which are not neutral and may introduce a point of view or attitude.
\end{itemize}

\section{Background and Related Work}

Prior work falls into following main areas:  online community norms, writing style norms, norm enforcement, effects of norm enforcement, and motivation for participation and editor roles in Wikipedia.

\subsection{Online Community Norms}

Norms are habitual behaviors that characterize a social group and differentiate it from other social groups~\cite{hogg2006social}. 
While norms are explicitly defined through detailed guidelines (e.g., Reddit) and FAQs (e.g., Usenet) in some communities, in other communities norms are not formally codified, but emerge socially through the interactions of the members~\cite{burnett2003beyond}. Online communities exhibit norms~\cite{fiesler2018reddit} related to member behavior such as content appropriateness~\cite{bergstrom2011don}, writing style~\cite{wiki2018style}, and adherence to community policies~\cite{lampe2004slash}. Community norms evolve as its members negotiate norms~\cite{kim2000community} and as newcomers arrive and old members become less active~\cite{danescu2013no}. While community norms are studied in terms of deviant behavior such as abusive language~\cite{sood2012profanity}, trolling~\cite{cheng2015antisocial}, and vandalism~\cite{geiger2010work}, norms also encourage participation~\cite{ren2012building} and help communities achieve their goals~\cite{bryant2005becoming, chancellor2018norms}. As a collaborative content production community, Wikipedia has several norms~\cite{wiki2018norms} related to neutral point of view~\cite{wiki2018npov}, supportive communication~\cite{reagle2010nice}, vandalism~\cite{priedhorsky2007creating}, and member participation~\cite{bryant2005becoming}. In this work, we focus on Wikipedia's norms related to NPOV writing style. 

\subsection{Writing Style Norms}
%
\emph{Communicative competence} is the ability to use language appropriately, in accord with social situation and associated norms~\cite{hymes1972communicative}. This notion of communicative competence can be used to explain linguistic norms of communities---the language used by community members should not only be correct, but should also be appropriate. 
A related concept is \emph{legitimate language}, which is defined in any community as the language produced by a subset of speakers/writers with symbolic authority, and is often codified into explicit standards~\cite{bourdieu1991language}. In the context of collaborative online writing, legitimate language can be considered to be the language produced by the high-status members of the community, and is governed by community policies.

Among online communities, Wikipedia is well known for its focus on the quality of the content produced by volunteer contributors~\cite{bryant2005becoming}. Wikipedia provides detailed guidelines of the preferred writing styles through a manual of style~\cite{wiki2018style}. A large body of research focused on Wikipedia's writing style norms including the prediction of article quality~\cite{lipka2010identifying}, biased content~\cite{al2012automatic}, quality flaws~\cite{anderka2012predicting,anderka2012breakdown}, and vandalism detection~\cite{harpalani2011language}.
Focusing specifically on bias in Wikipedia, \citet{recasens2013linguistic}  built a dataset of Wikipedia phrases that are corrected for NPOV, and then trained a classifier using linguistic features from hand-crafted lexicons of factive verbs, hedges, and subjective intensifiers. We build on this prior work by using these lexicons to characterize biased language, but note that while \citet{recasens2013linguistic} aimed to identify the linguistic aspects of norms, our focus is to understand the effects of NPOV tagging and correction. Specifically, we are interested in the change in the rate of biased language use when an article has an active NPOV tag, and when an editor is corrected for NPOV.

\subsection{Enforcing Norms}

Deviant behavior is an ongoing challenge to online communities, requiring persistent regulatory effort~\cite{kiesler2012regulating}. 
In addition to providing detailed guidelines, online communities also take active moderation actions against the violations of community guidelines~\cite{buzzfeed2015redditban}, including both technical and social approaches~\cite{grimmelmann2015virtues}. Technical moderation includes banning posts with any keywords from predefined word lists (e.g., Yik Yak~\cite{draper2018distributed}) and banning posts based on the IP addresses from which posts originate (e.g., Hacker News~\cite{hackernews2018ipban}, Yelp~\cite{rahman2014turning}). Social moderation includes both distributed approaches~\cite{lampe2014crowdsourcing} where the community members vote on content (e.g., Yik Yak, Slashdot)~\cite{lampe2004slash} and centralized approaches where a small number of users called moderators maintain community practices by removing posts they find inappropriate (e.g., Reddit) \cite{grimmelmann2015virtues}.

In Wikipedia, a rich set of policies and guidelines articulate strategies for seeking consensus,  principles of encyclopedic content, and appropriate user behavior. The policy environment in Wikipedia encodes and explains norms, but the policies are not imposed top down; rather policies are created and managed by the Wikipedia community itself~\cite{beschastnikh2008wikipedian}.  Norms are enforced by policy citations~\cite{beschastnikh2008wikipedian}, where template messages~\cite{di2008wiki} are added to articles and discussion talk pages~\cite{jan2017testing}. The policy pages are subject to the same editing processes of Wikipedia articles and contributors are given a participatory role when creating or editing policy pages. While several quality control mechanisms to enforce objective language in Wikipedia articles have been in place for a long time, the impact of these mechanisms has not been quantitatively measured. This gap in the literature motivates our broad research question in this study: whether NPOV tagging and corrections help Wikipedia articles and editors to converge to desired writing styles norms.

\subsection{Effects of Norm Enforcement}
Most prior work on the effects of norm enforcement has focused on moderation techniques related to the design of online communities. Users report significantly higher intent to participate in a community that is moderated, compared to an unmoderated community~\cite{wise2006moderation}.
A study on Slashdot --- a popular website with a good amount of moderation --- found that users come to a consensus about the community's moderation policies and such moderation practices enable large scale civil participation~\cite{lampe2014crowdsourcing}. Reddit's decision to ban hate communities resulted in a reduction of abusive language~\cite{chandrasekharan2017you}.
An analysis of different moderation styles in online health support communities found that positive and rewarding moderation styles are more effective than negative and punishing styles~\cite{matzat2014styles}. In online discussion forums related to education, peer moderation is found to be more encouraging active participation than moderation by superiors~\cite{seo2007utilizing}.

The effects of moderation on the quality of collaboratively created content and long term behavior of individual community members have not been investigated much. In the context of Wikipedia, \citet{halfaker2011don} found that the action of ``reverting'' edits has the effect of reducing motivation and quantity of work, particularly for new editors. However, they also found that reverts result in higher quality contributions. \citet{halfaker2013rise} found that enforcing quality control mechanisms affects the retention of high-quality newcomers. Personalized warning messages, as opposed to pre-defined template messages, are found to be effective in retaining newcomers~\cite{geiger2012defense}. 
Motivated by this prior literature, which studies the effectiveness of norm enforcement actions both at the platform-level (i.e., articles) and individual member-level (i.e., editors), we seek to answer our research question about the effectiveness of NPOV norm enforcement both at the article level (RQ1) and at the editor level (RQ2). 


\subsection{Motivation for Participation and Editor Roles in Wikipedia} \label{sec:editor roles}
The scale and success of Wikipedia as a volunteer-run collaborative content producing community have attracted a large body of research  about various aspects of Wikipedia including motivations to contribute~\cite{forte2006wikipedia, nov2007motivates, schroer2009voluntary}, formation of community~\cite{benkler2006wealth, ciffolilli2003phantom, voss2005measuring}, content quality~\cite{lih2004wikipedia,chesney2006empirical}, editorial and authorship attributes~\cite{emigh2005collaborative}, sharing and collaborative knowledge building~\cite{lih2004wikipedia}, learning~\cite{forte2006wikipedia}, coordinations of complex tasks~\cite{keegan2013hot}, and editor functional roles~\cite{arazy2015functional, forte2005people}. Prior studies on the motivations for voluntary contribution to Wikipedia provide several explanations including the ideology for contributing to Wikipedia as a variant of open-source application~\cite{nov2007motivates}, internal self-concept motivation~\cite{yang2010motivations}, cognitive (e.g., learning new things or intellectual challenge) and affective reasons (e.g., pleasure)~\cite{rafaeli2008online}, and the sense of individual efficacy~\cite{bryant2005becoming}.

Another line of work focused on different editor roles that emerge in Wikipedia~\cite{bryant2005becoming, kriplean2008articulations, panciera2009wikipedians, welser2011finding, arazy2015functional}. \citet{bryant2005becoming} find that while novice editors contribute edits to articles related to their domain of expertise, expert editors contribute towards improving the quality of Wikipedia itself. \citet{arazy2015functional} identified functional roles in Wikipedia such as technical administrator, border patrol, quality assurance technicians, administrators, and directors, and find that editor roles determine activity patterns across variety of editing tasks. \citet{yang2016did} automatically identify various editor roles based on low-level edit types and find that different editor roles contribute differently in terms of edit categories and articles in different quality stages require different types of editors. In this work, we focus on editor roles that make contributions to improve the quality of Wikipedia articles, particularly articles in the NPOV dispute category.

\section{Datasets}
\begin{figure}[h]

\centering
\includegraphics[width=0.95\textwidth]{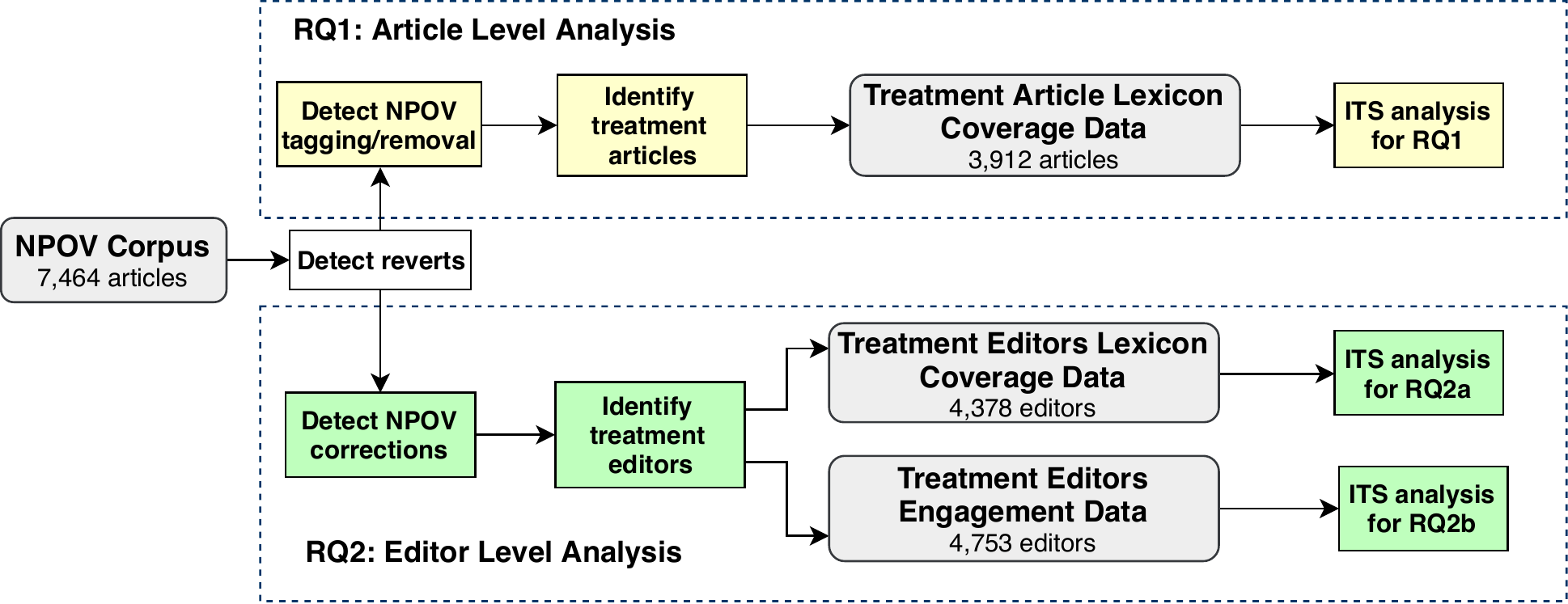}
\caption{Flowchart showing the data and method setup for each research question.}
\label{fig:flowchart}
\end{figure}

In this section we describe the datasets and extraction of treatment articles and editors. \autoref{fig:flowchart} summarizes our data pipeline and methods.

\subsection{Wikipedia NPOV Corpus} \label{npov-corpus}
We begin with the NPOV corpus from ~\citet{recasens2013linguistic}, which they used to build a classifier to detect bias inducing terms in phrases. The NPOV corpus contains  7,464 articles from the English Wikipedia with a total of  2.7 million revisions. 
It was constructed by retrieving all the articles that were in the \textit{NPOV dispute} category\footnote{\url{https://en.wikipedia.org/wiki/Category:All_NPOV_disputes}} in early 2013, together with their full revision history. Wikipedia editors are encouraged to identify and revise biased content to achieve neutral tone, and several NPOV tags are used to mark biased content.\footnote{e.g., \texttt{\{\{POV\}\}}, \texttt{\{\{POV-check\}\}}, \texttt{\{\{POV-section\}\}}, etc. When such tag is added, a template such as the following is displayed in the article page: ``The neutrality of this article is disputed. Relevant discussion may be found on the talk page. Please do not remove this message until the dispute is resolved.''} Articles that are marked with any of the NPOV tags will be listed in Wikipedia's category of \textit{NPOV disputes}. Each version of the article is considered as a \textit{revision}, and an article is a set of chronologically ordered revisions. Each revision of the article contains the version of the article content along with metadata such revision ID, timestamp, contributor, revision comment, and an SHA-1 hash key.


\subsection{Identifying NPOV Tagging and Removal}
\label{sec:identify-npov-tags}
Several NPOV tags are used to mark biased content and there is no standardized template or metadata available to directly detect tag addition and removal. Therefore, to identify in which revision an article was tagged as violating NPOV and when that tag was removed, we used a set of NPOV-tag patterns based on a preliminary inspection of a small subset of NPOV tagged articles.\footnote{\texttt{\{\{.*(POV|Pov|PoV|Point Of View|pov|NPOV|Npov|npov|NEUTRALITY|Neutrality|neutrality\\|NEUTRAL|Neutral|neutral).*\}\}.}} 
In order to detect the addition or removal of an NPOV tag in the absence of any readily available metadata, we need to compare each consecutive revisions and then inspect the difference in the textual content for an addition or removal of an NPOV tag. To automatically extract the changes between two consecutive revisions in terms of added and removed content, we used the Diff Match and Patch library,\footnote{\url{http://code.google.com/p/google-diff-match-patch}} which was also used by \citet{recasens2013linguistic}. If any of the NPOV tags was present in the added content, then we mark that revisions as an NPOV tag addition point for that article. Similarly, if any of the NPOV tags was present in the removed content, we mark that revision as an NPOV tag removal point for that article. 

Accurate identification of valid tag additions and removal is challenging due to the presence of vandalism and reverts.\footnote{Vandalism and edit wars are primary challenges for Wikipedia editors. There is a large body of work regarding this. For example, see \citet{geiger2010work, shachaf2010beyond, potthast2008automatic}.} From our initial inspections of a small set of NPOV tagged articles, we identified several cases of vandalism (e.g., NPOV tags were removed intentionally or accidentally, and then added back within the next few revisions; revisions containing NPOV tags were reverted to a prior revision without a tag). To minimize the effect of vandalism in the accurate identification of NPOV tag addition and removal, we used two heuristics. First, we used the SHA-1 hash key present in the revision metadata to track reverts and disregarded reverted revisions. 
Second, if an NPOV tag was added immediately after a prior tag removal, we consider the last tag removal as unreliable. If such changes appear within five revisions after a tag removal, we merge the consecutive tag addition-removal pairs and consider only the latest revision as a valid tag removal revision. In this way, we identified 6,512 articles that had at least one NPOV tag addition and 1,095 articles with multiple NPOV tag additions.

\subsection{Identifying Treatment Editors}
\label{sec:identify-treatment-editors}

For our editor-level analysis, we define treatment as editing part(s) of an article's content to correct for NPOV. Treatment editors are editors who originally contributed the portion of the text that was corrected for NPOV. To identify whether a revision was corrected for NPOV, we first check if the comment metadata for that revision contains ``NPOV'' or ``POV'' or any case variations. If one of these strings appear in the comment and if that revision was not identified as an NPOV tag addition or removal revision, then we consider it to be a \textit{NPOV correcting} revision. While \citet{recasens2013linguistic} checked for only the presence of NPOV strings in the comment to detect NPOV corrections, in our initial inspection we found that sometimes comments of NPOV tagging and removal revisions also contain these NPOV strings. Therefore, we additionally checked if the revision had a tag addition or removal as we do not consider those as NPOV correcting revisions.\footnote{Note that an NPOV removal revision could include corrections for NPOV and removal of NPOV tag. However, in our initial inspections, we found that NPOV issues are often solved in revisions prior to the NPOV tag removal revision. In this way, few NPOV correcting instances could have been missed, but we aimed for precision rather than recall.}

Once we have identified NPOV correcting revisions, we trace back previous revisions to identify the treatment editors---editors who originally contributed the portion of text that was corrected for NPOV. As a preprocessing step, we removed revisions which are reverts of previous revisions in order to identify the editor who originally contributed a span of text. Otherwise, the editor who made the revert could have been attributed to a portion of text. To trace the treatment editors, we first built a detailed record of which part of the article was contributed by which editor at the character granularity level from the beginning of the article revision history. We then checked all of the NPOV correcting revisions, extracted their first \textit{diff} compared to the previous revision, and looked into the record to identify which editor originally contributed to that \textit{diff-ed} language.\footnote{Note that recent work such as \citet{flock2014wikiwho} proposed other sophisticated algorithms to attribute authorship of revisioned content. Future work could consider employing such algorithms. }


Automated scripts or bots are used in Wikipedia to perform repetitive and mundane tasks such as minor fix for syntax and adding dates to NPOV tags. Because our interest is in studying the writing style of human editors, we removed bots by checking for the presence of ``bot'' in the username and presence for the ``bot'' editor group attributes.\footnote{Editor attributes are obtained from the Media Wiki API~\url{https://www.mediawiki.org/wiki/API:Query}. Examples of editor group attributes include ``reviewer'', ``administrator'', and ``bot''.} In total, we identified 5,645 treatment editors after removing 20 bot accounts. Note that each treatment editor could be NPOV corrected more than once. On average, a treatment editor is corrected for NPOV 1.9 times, and 32.4\% of the treatment editors are corrected for NPOV more than once.

\subsection{Dataset for Article-Level Analysis} \label{article-level-data}
In RQ1 we study the effect of NPOV tagging on the linguistic style of the articles. For this analysis we use the revisions of articles in the NPOV corpus (\cref{npov-corpus}) and characterize the linguistic style of each of the revisions. To observe the biased language trend over time, we limited this dataset to include NPOV articles which have at least 40 revisions after the treatment (i.e., addition of an NPOV tag) and considered up to 40 revisions before and after the treatment for the regression analysis.\footnote{We also considered the number of revision threshold window $W$ of $[10 \le W < 20]$ and $[20 \le W < 40]$ and the results are similar qualitatively. The number of articles included in the analysis for each of these thresholds are 1054, 1095, and 1763 for the revision windows $[10 \le W < 20]$, $[20 \le W < 40]$, and $[W \ge 40]$ respectively.  }

\subsection{Dataset for Editor-Level Analysis} \label{user-level-data}
To study the editor-level effects of NPOV correction, we collected each treatment editor's textual contribution to the Wikipedia article namespace\footnote{Wikipedia includes multiple namespaces such as article page, article talk page, user page, user talk page, etc. Since our interest in to characterize editor writing style in Wikipedia articles, we restricted our dataset to the article namespace. Prior work, such as \citet{elia2006analysis}, has found significant differences between the linguistic style of article text and talk page text due to the conversational nature of talk pages.} by querying the Media Wiki API.\footnote{\url{https://www.mediawiki.org/wiki/API:Query}} In RQ2a we study the effect of NPOV correction on the writing style of the treatment editors. To characterize treatment user writing style before and after treatment, we first obtained up to 150 revisions to which the editors contribute (in all Wikipedia articles, not restricting to the initial NPOV corpus) before and after their first treatment point. To control for any platform level changes such as revisions to NPOV policies, we restricted the pre and post treatment revisions to be within an year from the treatment. We then queried for the parent revision of each of the editors' revisions and extracted the text added by the treatment editors using the \textit{diff} library. Since our interest is in the language use, we considered only the revisions with any content addition.\footnote{Examples of other revision contributions include addition of links, templates, and citations.} In RQ2b we study the changes in treatment editor engagement before and after NPOV correction. To characterize treatment editor engagement we obtained all the revisions (in all Wikipedia articles, not restricting to the initial NPOV corpus) they have contributed to during two months before and after the treatment point. 
   
\paragraph{Attributing Editor Content Contribution.}
Our approach to identify the editor who originally contributed NPOV corrected content uses the \textit{diff} of two consecutive revisions as described in \Cref{sec:identify-treatment-editors}. One potential issue with this approach is content mis-attribution due to vandalism~\cite{de2013attributing}. To minimize this issue, we identify and remove reverted revisions. 
When characterizing editor language (\Cref{{user-level-data}}), we compare each of the editor's revision to its parent revision and compute the \textit{diff} of both revisions. When making revisions, editors could re-introduce content that was previously contributed by another editor. Even though the re-introduced content may not be originally contributed by the editor who makes the revision, we assume that the editor is at least partially responsible for the language used in that content as they include it in the revision they make. 
While our approach is simple in terms of implementation, other sophisticated approaches (e.g., \citet{flock2014wikiwho}) exist and such approaches are specifically designed for the task of authorship attribution for revisioned content. We acknowledge that such methods are superior to our approach, and future work could use an implementation of such algorithms.

\section{Methods}
In this section we describe methods related to preprocessing of Wikipedia data for our analyses, identifying biased language, and measuring effect of treatment.

\subsection{Identifying Biased Language} \label{subsec:lexicons}
Bias in Wikipedia can emerge from several factors including language style, editors' point of view, cited sources, coverage of topics, etc. Prior work focused on political bias~\cite{greenstein2012collective}, cultural bias~\cite{callahan2011cultural}, gender bias~\cite{wagner2015s}, topic bias~\cite{ferschke2013impact}, and authoritative bias~\cite{das2016manipulation}. Our goal in RQ1 and RQ2a is to study the effects of NPOV tagging in article-level and editor-level language. Particularly, we are interested in characterizing the bias in articles and editor contribution in terms of the linguistic style that may introduce bias. The Wikipedia Manual of Style~\cite{wiki2018style} states that ``\textit{There are no forbidden words or expressions on Wikipedia, but certain expressions should be used with caution, because they may introduce bias. Strive to eliminate expressions that are flattering, disparaging, vague, or endorsing of a particular view point.}'' Prior work \cite[e.g.,][]{herzig2011annotation, recasens2013linguistic} has addressed the task of identifying biased language in Wikipedia at different levels. However, we are not aware of a reliable system that can detect bias in terms of linguistic style in Wikipedia articles. Therefore, we use a set of linguistic style lexicons as a proxy to characterize biased language in Wikipedia.\footnote{We limit our analysis only to lexicon, while syntax, semantics, and pragmatic measures could also be studied.} 

The Wikipedia Manual of Style provides a list of \textit{words to watch} in a prescriptive manner~\cite{wiki2018wikiwords}, indicating style words that may introduce bias. We compiled these styles words into a lexicon called \textit{Words to Watch}. Prior work of \citet{recasens2013linguistic} introduced the task of detecting bias inducing terms in phrases from Wikipedia articles and used a set of pre-compiled style lexicons that are indicative of expressions of attitude or point of view. These lexicons include hedges, factive verbs, assertive verbs, positive words, and negative words. We use these nine lexicons in addition to the words from Wikipedia manual.\footnote{Note that \citet{recasens2013linguistic} used a bias lexicon created from the NPOV articles. We do not include it for our analysis as this lexicon is not entirely indicative of linguistic style and included content words such as \textit{communist}, \textit{historian}, and \textit{migration}.} A list of all the lexicons we used is shown in \autoref{tab:bias-lexicons} with a description, sources, and example terms. To characterize the amount of biased language in a text, we compute the coverage of each lexicon words per token in the text. 

\begin{table}%
\caption{List of Lexicons Used to Characterize Bias Language.}
\label{tab:bias-lexicons}
\begin{minipage}{\columnwidth}
\begin{center}
\begin{tabular}{@{}m{.18\textwidth}m{.50\textwidth}m{.20\textwidth}@{}}
\textbf{Lexicon Name} & \textbf{Description} & \textbf{Example Terms} \\
  \midrule
{\textbf{Words to Watch }}&  Style words that may introduce bias (\citet{wiki2018wikiwords})& \textit{fortunately}, \textit{notable}, \textit{often}, \textit{speculate}\\[6pt]
\textbf{Hedges}  & Used to reduce one's commitment to the truth of a proposition (\citet{hyland2005metadiscourse}) & \textit{apparent}, \textit{seems}, \textit{unclear}, \textit{would}\\
\textbf{Assertives} & Complement clauses that assert a proposition (\citet{hooper1974assertive})&\textit{allege}, \textit{hypothesize}, \textit{verify}, \textit{claim}\\
\textbf{Positive Words }&Positive sentiment terms (\citet{liu2005opinion})&\textit{achieve}, \textit{inspire}, \textit{joyful}, \textit{super}\\
\textbf{Negative Words}&Negative sentiment terms (\citet{liu2005opinion})&\textit{criticize}, \textit{foolish}, \textit{hectic}, \textit{weak}\\
\textbf{Factives}&Terms that presuppose the truth of their complement clause (\citet{kiparsky1970fact})&\textit{regret}, \textit{amuse}, \textit{strange}, \textit{odd}\\
\textbf{Implicatives}&Imply the truth or untruth of their complement, depending on the polarity of the main predicate (\citet{karttunen1971implicative})&\textit{avoid}, \textit{hesitate}, \textit{refrain}, \textit{attempt}\\
\textbf{Report Verbs}&Used to indicate that discourse is being quoted or paraphrased (\citet{recasens2013linguistic})&\textit{praise}, \textit{claim}, \textit{dispute}, \textit{feel}\\
\textbf{Strong  ~~~~~~~~~~~~~~~~~Subjectives}& Add strong subjective force to the meaning of a phrase (\citet{riloff2003learning})&\textit{celebrate}, \textit{dishonor}, \textit{overkill},\textit{worsen}\\
\textbf{Weak  ~~~~~~~~~~~~~~~~~Subjectives}&Add weak subjective force to the meaning of a phrase (\citet{riloff2003learning})&\textit{widely}, \textit{unstable}, \textit{although}, \textit{innocently} \\
\end{tabular}
\end{center}
\bigskip\centering
\end{minipage}
\end{table}

\subsection{Measuring the Effect of Treatment on Biased Language Usage}
Our goal in RQ1 is to investigate whether NPOV tag addition has an effect on the level of biased language in the NPOV tagged articles. Similarly, our goal in RQ2a is to investigate whether NPOV correction has an effect on the level of biased language usage of treatment users. Both of these are causal questions. To answer them, we apply the Interrupted Time Series technique for causal inference~\cite{bernal2017interrupted}. This method has been used in recent studies on hate speech~\cite{chandrasekharan2017you} and conspiratorial discussions\cite{samory2018conspiracies}.
Interrupted Time Series analysis is a quasi-experimental design that can be used to evaluate the longitudinal effects of an intervention (i.e., treatment), through segmented regression modeling. The term quasi-experimental refers to an absence of randomization. ITS is a tool for analyzing observational data where complete randomization, or case-control design, is not affordable or possible. A notable strength of ITS with respect to measuring treatment effects using observational data is that ITS controls for the effects of secular trends in a time series of outcome measure. For example, if a measure is increasing or decreasing historically over time, and if we only look at the averages pre- and post- treatment, we would not notice the change in slope or level.

In its basic form, an ITS is modeled using a regression model (e.g., linear, logistic or Poisson) that includes only three time-based covariates as shown in \autoref{eqn:its-model}. The regression coefficients of these covariates estimate the pre-treatment trend, the level change at the treatment point, and the trend change from pre-treatment to post-treatment. 

\begin{equation}
\label{eqn:its-model}
Y_{t} = \beta_{0} + \beta_{1}T + \beta_{2}X_{t} + \beta_{3}TX_{t}
\end{equation}

Here $T$ is the time elapsed since the start of the study; $X_{t}$ is a dummy variable indicating the pre-treatment period (coded 0) or the post-treatment period (coded 1); $Y_{t}$ is the outcome at time $t$.
The pre-treatment slope, $\beta_{1}$, quantifies the trend for the outcome before the treatment; the level change, $\beta_{2}$, estimates the change in level that can be attributed to the treatment; the change in slope, $\beta_{3}$, quantifies the difference between the pre-treatment and post-treatment slopes. This model of ITS assumes that without the treatment, the pre-treatment trend would continue unchanged into the post-treatment period, and there are no external factors systematically affecting the trends. One important step before modeling an ITS analysis is to hypothesize how the treatment would impact the outcome if it were effective, particularly whether the change will be in the slope of the trend, a change in the level, or both. 

While we assume the events of NPOV tagging and subsequent NPOV corrections to be independent, editors could tag an article for NPOV as a preparation for their subsequent revisions. In such cases, NPOV tagging could be a precursor to future revisions by the same editor. Therefore, in our analysis for RQ1, we remove all the articles where the editor who tagged an article for NPOV contributed a revision in the subsequent 40 revisions after the NPOV tagging revision.\footnote{We removed 47 articles in this way, which is nearly 2.7\% of the total number of articles considered for RQ1. }

\section{RQ1: Article-Level Effects of NPOV Tagging}


\subsection{Article Writing Style: RQ1}
\subsubsection{Quantifying Article Writing Style} 
We use the dataset we created for the article level analysis (\cref{article-level-data}) to study RQ1. Before measuring the writing style of each article revision, we performed a set of preprocessing steps to extract textual contribution of the article revision. We removed Wikipedia markup templates, URLs, and other non-textual content. We then tokenized the text and extracted tokens with two or more characters. To quantify article writing style, we extracted the coverage of tokens from each of the ten bias-related lexicons (\cref{subsec:lexicons}) and computed lexicon coverage rate as the number of lexicon tokens per total tokens in the text of each article revision. We considered the latest 40 pre-treatment revisions and the earliest 40 post-treatment revisions of the NPOV articles for the ITS analysis. 

\subsubsection{ITS Modeling for Article Writing Style}
When an article has an active NPOV tag, it stays on top of the article for all revisions until its removal (\autoref{fig:npov-tag}). As this can be considered a continuous treatment, we would expect the editors to be cautious for biased language and the articles to have a continuous drop in biased language coverage. Therefore, to measure the effect of an article having an active NPOV tag on the biased language coverage, we hypothesize a slope change in the article-level biased language usage after treatment. To model this, we use the ITS regression model in \autoref{eqn:its-model}  and use a linear model to fit in the R software. 

\subsection{Results: RQ1}
\begin{figure}[h]
\centering
\includegraphics[width=0.98\textwidth]{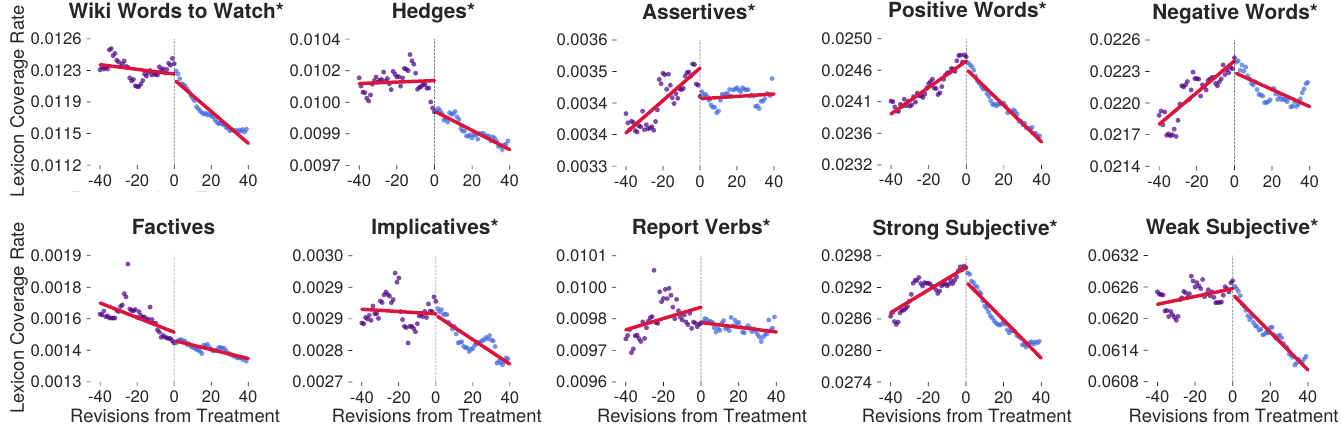}
\caption{Article level lexicon coverage with respect to the treatment of NPOV tagging. We computed the average lexicon coverage rate of treatment articles' 40 revisions before and after the treatment revision. Lexicons for which we observe a statistically significant drop in slope ($p < 0.05$) are marked with an asterisk (*).}
\label{fig:article-trend}
\end{figure}

Results of the ITS analysis for RQ1 is shown in \autoref{tab:rq1-results} and the trends are shown in \autoref{fig:article-trend}. Through the ITS analysis, we observe that when an active NPOV tag is on the article, there is a statistically significant change in the slope of the trend of nine out of ten lexicons used to characterize biased language. 
\paragraph{\textit{Robustness Check:}}
The dataset used in this analysis was collected from the snapshot of articles in Wikipedia's \textit{NPOV dispute} category in early 2013. Some of the articles could contain NPOV issues that are inherently difficult to resolve and could be tagged for a long period of time. As this could be a potential confound in the editing behavior of such articles, as a robustness check, we re-ran the analysis on a subset of the data after removing articles that are NPOV tagged for more than three years. The distribution of time-intervals between the addition of the NPOV tag and the data collection time, and the result of the regression analysis for the subset of articles after removing potential outliers are shown in \hyperref[sec:appendix]{Appendix A}. The treatment effect remains significant after excluding these long-running disputes.

\begin{table}%
\caption{Results of RQ1 Analysis. Article lexicon coverage is computed for the textual content of the article results from each revision. Coefficient $\beta_{3}$ indicates the change in slope after treatment. Statistical significance after correcting for multiple comparisons using \citet{benjamini1995controlling} adjustment for false discovery rate are shown; *: $p < 0.05$, **: $p < 0.01$, ***: $p < 0.001$. Percentage change is computed as the change in post-treatment lexicon coverage after 40 revisions.}
\label{tab:rq1-results}
\begin{minipage}{\columnwidth}
\begin{center}
\begin{tabular}{lrrlr}
  \textbf{Lexicon} & \textbf{$\beta_{3}$}  & \textbf{$p$-value} &  & \textbf{\% change}\\
\midrule
Words to Watch&-1.51e-05&1.76e-13&*** &-5.44\\
Hedges&-5.75e-06&2.76e-08&*** &-3.92\\
Assertives&-3.57e-06&1.89e-10&*** &-6.10\\
Positive Words&-4.48e-05&< 2e-16&*** &-7.72\\
Negative Words&-2.31e-05&< 2e-16&*** &-4.62\\
Factives&1.35e-06&8.76e-01& &0.90\\
Implicatives&-2.63e-06&1.50e-05&*** &-3.79\\
Report Verbs&-3.18e-06&1.27e-03&** &-1.90\\
Strong Subjectives&-5.78e-05&< 2e-16&*** &-8.74\\
Weak Subjectives&-4.34e-05&< 2e-16&*** &-2.95\\
\end{tabular}
\end{center}
\bigskip\centering
\end{minipage}
\end{table}%

\section{RQ2: Editor-Level Effects of NPOV Correction}
Next, we explore the editor-level effects of NPOV correction through the following research questions:
\begin{itemize}
 \item[]\indent \textbf{{RQ2:} } \textbf{What are the editor-level effects of NPOV correction?}
 \item[]\indent RQ2a: {What effect does NPOV correction have on the writing style of the treatment editors?}
 \item[]\indent RQ2b: {What effect does NPOV correction have on the engagement of treatment editors?}
\end{itemize}

\subsection{Editor Writing Style: RQ2a}
\subsubsection{Quantifying Editor Writing Style} 
We use the dataset we created for the editor level analysis (\cref{user-level-data}) to study RQ2a. Before measuring editor writing style, we performed a set of preprocessing steps to extract textual contribution of the treatment editors. We removed Wikipedia markup templates, URLs, and other non-textual content. We then tokenized the text and extracted tokens with two or more characters. To quantify editor writing style, we extracted the coverage of tokens from each of the ten bias-related lexicons (\cref{subsec:lexicons}) and computed lexicon coverage rate as the number of lexicon tokens per total tokens in the text contributed by an editor in a revision. We considered the latest 40 pre-treatment edits and the earliest 40 post-treatment edits of the treatment editors for the ITS analysis.

\subsubsection{Controlling for Editor Experience and Talk Page Discussion}
Other factors may confound measurement of the effect of NPOV correction on editor-level writing. One such factor is the experience of the editor: previous work shows that inexperienced editors respond differently to norm-enforcement activities such as edit reverts~(e.g., \citep{halfaker2011don}).
To control for editor experience, we include it as a binary predictor in the regression model. Based on the criteria used in prior work~\cite{panciera2009wikipedians}, we consider editors who have contributed at least 250 edits prior to the treatment as \textit{experienced editors} and others as \textit{inexperienced editors}. In our dataset, 2,847 out of 4,378 editors had at least 250 edits prior to the treatment. 

In addition to correcting content for NPOV, editors sometimes also post to the talkpage of the articles or the talkpage of the editor whose contribution was revised to discuss about the correction. These posts trigger an email prompt to the original editor. Thus, the talk page discussion acts as an additional ``correction'' method that could potentially amplify the treatment of correcting content with NPOV tags in the comment metadata. The discussion between the correcting and corrected editors during the treatment period could influence the future behavior of the corrected editor in terms of language usage and engagement. Therefore, we also include the presence of talk page discussion as an additional binary predictor, which indicates whether any of the following criteria is satisfied during a week prior or after the treatment: (1) correcting editor posts on corrected editor's talk page; or (2) corrected editor posts on correcting editor's talk page; or (3) both corrected editor and correcting editor post on the article talk page. In our dataset, 1,216 of the 4,378 NPOV corrections were accompanied by a talk page discussion.

\subsubsection{ITS Modeling for Editor Writing Style}
Unlike an active NPOV tag on the article, which stays as a continuous treatment for all later revisions until its removal, the editor who gets an NPOV correction only receives treatment once.\footnote{While 32.4\% of the treatment editors are corrected more than once, repeated corrections are rare in the short window of 40 edits adjacent to the treatment.} %
 Therefore, we hypothesize a change in the \emph{level} of the trend of average editor writing style after treatment. We use the ITS regression model in \autoref{eqn:its-model}, excluding the term $\beta_{3}TX_{t}$ (since we are only hypothesizing a level change) and using a linear model fit in the R software (\emph{Model-I}). In addition to the terms in the ITS equation, as discussed in the previous subsection, we also used a second model (\emph{Model-II}) including two binary control predictors for editor experience ($experienced$) and talk page discussion during treatment ($discussion$). We also add an interaction term between each of these additional predictors and the post-treatment indicator (i.e., interaction with dummy term $X_t$).

\subsection{Results: RQ2a}
\begin{figure}[h]

\centering
\includegraphics[width=0.98\textwidth]{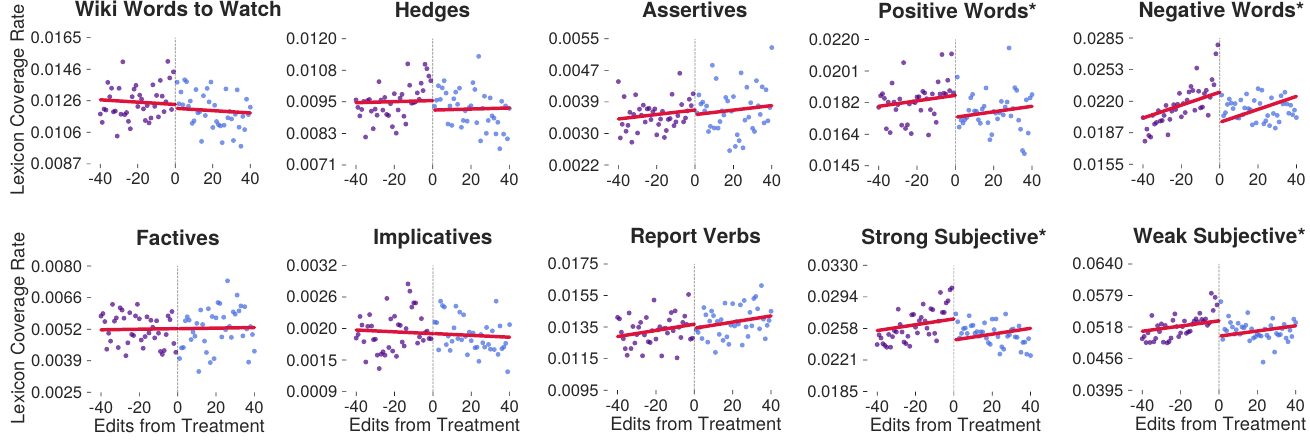}
\caption{Editor level lexicon coverage with respect to the treatment of NPOV correction. We computed the average lexicon coverage rate of treatment users' textual contributions 40 revisions before and after the treatment revision. Lexicons for which we observe a statistically significant level drop ($p < 0.05$) are marked with an asterisk (*).}
\label{fig:editor-trend}
\end{figure}

To understand the trend of the change due to treatment, we first visualize the level change in average editor writing style from \emph{Model-I} in \autoref{fig:editor-trend}. NPOV correction is associated with a small level decrease in the usage of four out of ten lexicons used to characterize biased language: positive words, negative words, strong subjectives, and weak subjective. No significant treatment effect is observed for the remaining six lexicons. \emph{Model-II} includes the control predictors for editor experience and talk page discussion. Relevant output from \emph{Model-II} is shown in \autoref{tab:rq2a-results}. The trend of level change ($\beta_2$) remains the same after including the control predictors. The coefficients for the $exprienced$ predictor are negative and significant for eight of the ten lexicons, which suggests that experienced editors use lower amount of non-neutral language.\footnote{This finding also validates our choice of lexicons to characterize non-neutral language as we would expect experienced editors to better adhere to community writing style norms compared to newbies.} The coefficients for the effect of discussion after treatment ($D : X_t$) are not significant except for the negative words lexicon. None of the coefficients for the effect of editor experience after treatment ($E : X_t$) are significant, suggesting that editor experience does not influence the effect of NPOV correction on writing style.  

\begin{table}%
\caption{Results of RQ2 Analysis after controlling for editor experience and talk page discussion during treatment. Editor lexicon coverage is computed for their textual contribution in each revision. Coefficient $\beta_{2}$ indicates the change in level after treatment; $D$ indicates whether there was any talk page discussion during treatment (reference: $dicussed = 0 $); $E$ indicates whether the editor is considered experienced or not (reference: $exprienced = 0 $); $X_t$ is the indicator variable for post-treatment; $D : X_t$ and $E : X_t$ are the interaction terms. Statistical significance after correcting for multiple comparisons using \citet{benjamini1995controlling} adjustment for false discovery rate are shown; *: $p < 0.05$, **: $p < 0.01$, ***: $p < 0.001$. Percentage change is computed as the level change following the treatment. Note that none of the coefficients for the $E : X_t$ interaction term are significant at $p < 0.05$.}
\label{tab:rq2a-results}
\begin{minipage}{\columnwidth}
\begin{center}
\begin{tabular}{lrlrrlrlrl}
  \textbf{Lexicon} & \textbf{$\beta_2$}   &   & \textbf{\% change} & \textbf{$D : X_t $} & & \textbf{$E$} & & \textbf{$E : X_t$} & \\
  \midrule
Words to Watch&-4.98e-04&&-1.92&-5.90e-04&&-2.98e-03&***&6.72e-04&\\
Hedges&-8.76e-04&&-3.85&-7.78e-05&&-1.38e-03&***&6.59e-04&\\
Assertives&1.58e-04&&-3.32&-3.05e-04&&-5.63e-04&*&-2.46e-04&\\
Positive Words&-2.27e-03&**&-6.96&-1.70e-04&&-3.47e-03&***&1.48e-03&\\
Negative Words&-2.55e-03&**&-13.54&-1.92e-03&*&-3.66e-03&***&5.41e-05&\\
Factives&3.51e-04&&-0.10&-5.44e-04&&1.48e-03&***&-3.78e-04&\\
Implicatives&-2.78e-06&&0.16&-1.72e-04&&-6.44e-04&***&2.22e-04&\\
Report Verbs&3.27e-04&&-1.86&-7.11e-04&&6.02e-04&&-6.03e-04&\\
Strong Subjectives&-2.48e-03&**&-9.02&1.59e-04&&-5.99e-03&***&-7.43e-06&\\
Weak Subjectives&-2.70e-03&*&-5.63&-2.18e-04&&-7.20e-03&***&-1.88e-04&\\
\end{tabular}
\end{center}
\bigskip\centering
\end{minipage}
\end{table}%

\subsection{Editor Engagement: RQ2b}

\subsubsection{Quantifying Editor Engagement}
To study RQ2b, we define engagement of a Wikipedia editor as any non-minor revisions they made to the article pages. We use the editor-level engagement dataset we created (\cref{user-level-data}) for this purpose and considered all treatment editors who have contributed at least ten revisions during the two months prior to the treatment. 

\subsubsection{ITS Modeling for Editor Engagement}
Similar to the hypothesis for editor level lexicon coverage, we hypothesize a change in the level of editor engagement after treatment. We model this using the ITS regression model in \autoref{eqn:its-model} excluding the slope term $\beta_{3}TX_{t}$, but with editor fixed effects. Because the dependent variable in this model is a count variable (i.e., number of revisions per day), we would use a Poisson regression model. However, summary statistics of the dependent variable show an excessive amounts of zeros ($>$ 40\%) and over-dispersion (variance $\gg$ mean). Therefore we used zero-inflated negative binomial regression model, as implemented in the \textit{pscl} package\footnote{\url{https://cran.r-project.org/web/packages/pscl/pscl.pdf}} in R software. Similar to RQ2a, we include binary controls for $experienced$ and $discussion$. 

\subsection{Results: RQ2b}
\begin{figure}[h]

\centering
\includegraphics[width=0.50\textwidth]{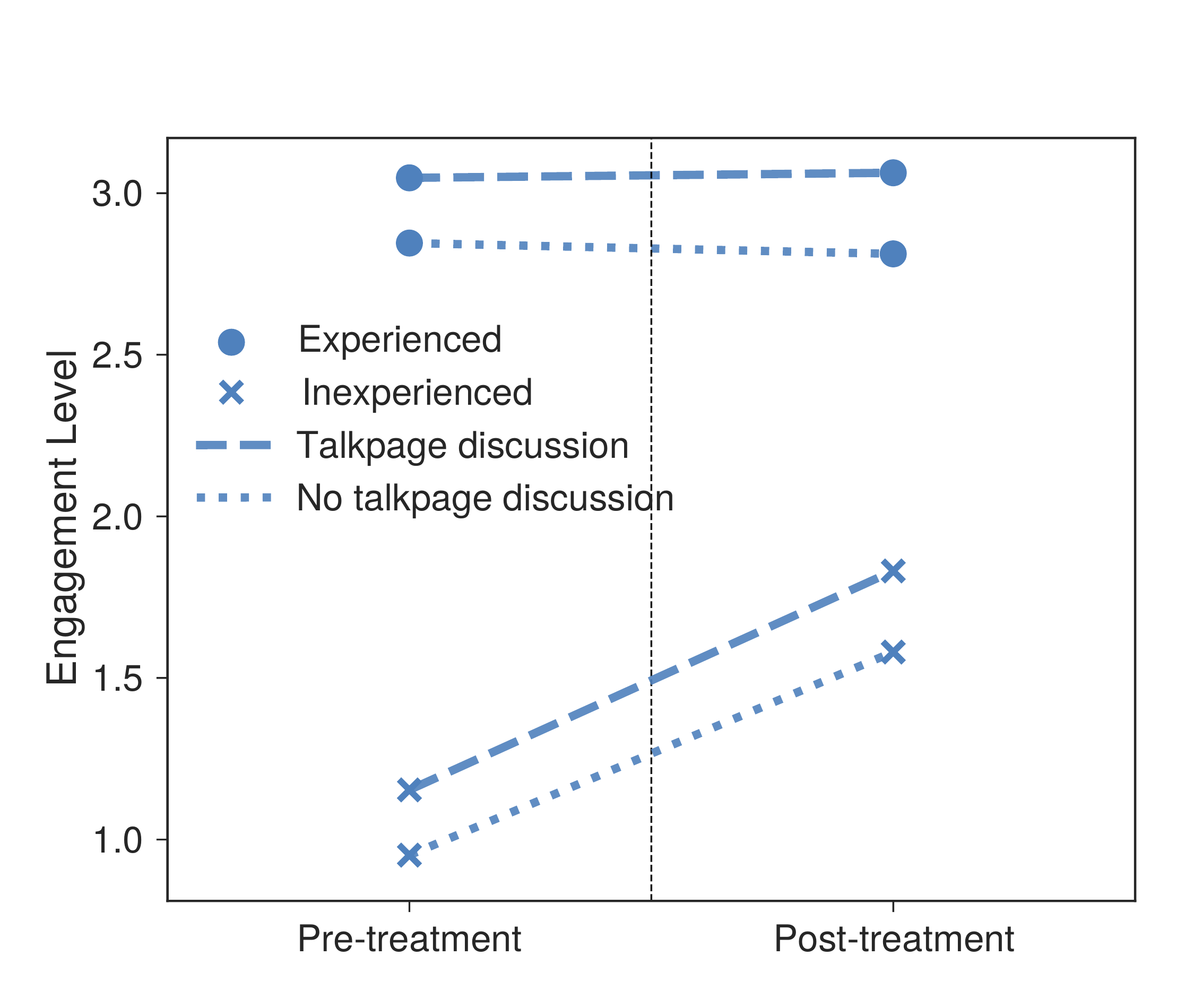}
\caption{Visualization of RQ2b regression coefficients showing the pre- and post-treatment engagement levels for the control factors $experienced$ and $discussion$.}
\label{fig:rq2b-cofficients-plot}
\end{figure}

Results of the ITS analysis for RQ2b are shown in \autoref{tab:rq2b-results}. The interaction between the level change and control predictors $experienced$ and $discussion$ are shown in \autoref{fig:rq2b-cofficients-plot}. These results show a significant increase in engagement for inexperienced editors (editors with less than 250 edits at treatment time). The decrease in engagement for experienced editors is small after including the overall temporal trend. Talk page discussion during treatment is associated with a small increase in engagement for both experienced and inexperienced editors.

\begin{table}%
\caption{Results of RQ2b Analysis: predictor coefficients of the count model of the zero-inflated negative binomial regression. The dependent variable, editor engagement is computed as number of edits during each of the three-day period. We considered 60 days prior to the treatment and 60 days after the treatment. $Post$-$treatment$ is the indicator for  whether the observation is post (=1) or pre (=0) treatment. $Timebin$ is the three-day window; we consider 20 timebin windows pre- and post-treatment, which includes 120 days in total. $Discussion$ is the indicator whether there was talk page discussion between correcting and corrected editors during treatment ($Dicussion=1$) or not. $Experienced$ is the indicator for whether the editor is experienced (>250 edits, $Experienced=1$) or not. Interaction terms include: Post-treatment : Discussion(=1) and Post-treatment : Experienced(=1). Statistical significance after correcting for multiple comparisons using \citet{benjamini1995controlling} adjustment for false discovery rate are shown;*: $p < 0.05$, **: $p < 0.01$, ***: $p < 0.001$. }
\label{tab:rq2b-results}
\begin{minipage}{\columnwidth}
\begin{center}
\begin{tabular}{lrrl}
  \textbf{Predictor} & \textbf{Coefficient}  & \textbf{$p$-value} &  \\
  \midrule
Post-treatment & 0.4675 &  <2e-16   & *** \\
Timebin & 0.0080&  <2e-16 & *** \\
Discussion(=1)      &   0.2018  & <2e-16  & *** \\
Experienced(=1)    &   1.8937  & <2e-16  & *** \\
Post-treatment : Discussion(=1) &  0.0491 & 0.0404  & *  \\
Post-treatment : Experienced(=1) & -0.6618  &   <2e-16  & *** \\
\end{tabular}
\end{center}
\bigskip\centering
\end{minipage}
\end{table}%

\section{Discussion}
Next, we reflect on the article-level and editor-level effects of NPOV tagging, discuss the differences in success for articles and editors, and conclude by considering implications of our findings to Wikipedia and other online communities with writing style norms.

\subsection{RQ1: Article-level Effects of NPOV Tagging}
When a Wikipedia article is tagged for not adhering to the community norm of neutral point of view, a banner template is displayed on the article stating that ``This article has multiple issues. Please help improve it or discuss these issues on the talk page.'' Articles marked with any of the NPOV tags are listed under Wikipedia's \textit{NPOV disputes} category and editors are encouraged to make edits to resolve the NPOV issues.
When an article has an active NPOV tag, we observe a statistically significant, gradual reduction in biased language coverage for all lexicons except factives.\footnote{Note that compared to other lexicons, the set of factives is small in size---containing only 27 terms---and hence the coverage of this lexicon is small.} The reduction is larger for Wikipedia words to watch (5.4\%), strong subjectives (8.7\%), positive words (7.7\%), and assertives (5.4\%), compared to other lexicons. Using these lexicons as a proxy for biased or non-NPOV language, it is evident that non-objective language used in Wikipedia articles decreases as editors make more revisions. This demonstrates the effectiveness of marking Wikipedia articles with NPOV tag in reducing the prevalence of non-neutral language. 

The dataset we used for this analysis is collected from a single snapshot of the articles in the \textit{NPOV dispute} category. Some of the articles which are controversial and hard to resolve could be NPOV tagged for a long period of time could affect the editing behavior. As a robustness check we repeated the analysis on the subset of the data after removing articles which has an active NPOV tag for more than three years. As we report in \hyperref[sec:appendix]{Appendix A}, the main results hold after removing these potential outliers.

\subsection{RQ2: Editor-level Effects of NPOV Correction}
Wikipedia's policies around NPOV language encourage editors to revise article content to correct for biased language. When editors correct non-neutral language, those revisions are logged with comments such as ``correcting for NPOV language'', ``removing POV'', and ``edits for NPOV''~\cite{recasens2013linguistic}. Using these comments, we extract revisions which are corrected for NPOV and using the revision history of the articles, we trace back to the editor who contributed the portion of text that was corrected for NPOV.

When an editor is corrected for NPOV, there is a significant reduction in their negative words usage (13.5\%); however, we do not observe any statistically significant change for six of the ten lexicons relating to biased language, including Wikipedia's own list of ``words to watch'' (RQ2a). These trends remain even after controlling for editor experience and talk page discussion during treatment. For RQ2b, we observe an increase in engagement for inexperienced editors after NPOV correction. One possible explanation is that NPOV correction helps inexperienced users become aware that their contributions are monitored by others, and this awareness could motivate them to contribute more.
These results for RQ2 suggest that when corrected for NPOV, the quality of the writing style of editors does not improve significantly (at least, as measured by all but one of our lexicons), while it leads to a increase in editing activities for inexperienced editors.
These findings partially conflict with the observations of 
\citet{halfaker2011don}, who found that revert actions demotivate new editors and reduce the quantity of work, even as they increase the overall quality of contributions.
\citet{halfaker2011don} also found that reverts affect editors differently based on the experience of the editor who makes the revert. Future work could perform similar analysis to further understand the effects of NPOV correction on engagement based on the experience of correcting editors.

\subsection{Possible Reasons for the Differences in  the Effectiveness of Norm Enforcement} \label{sec:tag-effectiveness}
While NPOV tagging helps articles to converge to the desired writing style, we do not find statistically significant evidence that these corrections encourage editors to adopt the desired writing style. One possible reason for the significant improvement in writing style of the NPOV tagged articles is the active contribution of a group of editors who are dedicated to improve the overall quality of Wikipedia articles. According to prior work on the classification of different types of work (refer \Cref{sec:editor roles}), editing tasks, and roles in Wikipedia, editors perform distinct roles, corresponding to distinct types of contributions. For example, \citet{bryant2005becoming} found that while novice editors contribute only to articles related to their expertise, expert editors contribute to improve the overall quality of Wikipedia articles. \citet{arazy2015functional} identified several roles of Wikipedia editors, including quality assurance technicians, who are editors contribute towards patrolling Wikipedia and ensuring content quality. Further, \citet{yang2016did} found that articles in different quality stages require different types of editors. Tools such as SuggestBot~\cite{cosley2007suggestbot} recommends specific editing tasks (e.g., cleanup and rewrite) to editors based on their previous editing patterns and interests. These prior findings support our hypothesis that the NPOV tagged articles are of interest to specific editor roles, who are dedicated to make revisions to improve NPOV tagged articles.

NPOV tags are highly visible to Wikipedia editors via the banner template displayed in the article (\autoref{fig:npov-tag}), often on the top, and the NPOV tagged articles are listed in a specific Wikipedia category (i.e., \textit{NPOV disputes} category). These visible actions could attract more editors to contribute towards improving the tagged articles. However, the treatment of NPOV correction of editor contribution is not visible to the same extent because editors are not directly notified about their contribution being corrected for NPOV. Editors keep track of the revisions to their previously contributed articles via the ``watchlist'' tool~\cite{bryant2005becoming}, but this requires active actions from the editors.


\subsection{Implications for Wikipedia and Online Language Moderation}

Our findings in RQ2a show that the current regime of tag-and-correct does not help editors to significantly improve their adherence to the NPOV norm. This suggests the need for additional interventions targeted at editors. Possible strategies include explicit notification of their textual contribution being corrected for NPOV language, reminders about writing style norms, and incentives when they progress. 
Findings from RQ2b shows an increase in engagement after treatment for inexperienced editors; talk page discussion during treatment is also assocaited with a small increase in engagement for all editors. The increase in engagement for inexperienced editors could be due to the signal that the community is aware of their contributions. Interventions such as explicit notification of NPOV correction could include personalized messages~\cite{geiger2012defense}\ that could further increase the engagement of inexperienced editors. 
Other online collaborative content creation communities who wish to enforce linguistic style norms may consider implementing interventions similar to Wikipedia's treatment of NPOV. Further community-level and individual-level interventions could help to converge to expected writing style norms. 


In this work, we focused on NPOV, which is one of the three core content policies of Wikipedia. The other policies are ``Verifiability'' and ``No original research''. These three policies jointly determine the quality of content acceptable in Wikipedia articles. Similar to NPOV tagging, these policies are enforced using tags such as \{\{citation needed\}\}, \{\{verification needed\}\}, \{\{original research\}\}, and \{\{synthesis\}\}. Our findings suggest that the NPOV tagging helps articles to converge to neutral language, but we did not find significant changes at individual editor language. As we discuss in \Cref{sec:tag-effectiveness}, various editor roles and the types of work performed by different editor roles could be one of the mechanisms by which NPOV tagging helps to improve article quality. A similar study could be performed to test the effectiveness of tagging for the other quality control policies of Wikipedia.

\section{Challenges, Limitations, and Future Work}

\begin{itemize}
  \item \textbf{Limitations with characterizing biased language.} We use a set of style lexicons including Wikipedia's ``words to watch'' to characterize non-neutral linguistic style. While similar approaches are used in prior work, these lexicons may not accurately and entirely capture non-neutral language. Machine learning could be applied to this problem by training on the edits that led to the application of an NPOV-related tag and correction~\citep{recasens2013linguistic}. We limit our analysis of language style to lexical methods; however, other aspects of language such as syntax, semantics, or pragmatics could also be studied (e.g., active vs. passive form).

  \item \textbf{Potential issues in accurately identifying the treatment groups.} The presence of reverts and vandalism imposed challenges in correctly identifying treatment articles and editors. We used a set of heuristics to reduce these issues (\cref{sec:identify-npov-tags}). However, our approach might not be perfect and could attribute an NPOV correction for an editor who is not the original contributor of the corrected text. This will result in an underestimation of treatment effects. Sophisticated algorithms to detect vandalism and reverts~\cite{priedhorsky2007creating, geiger2010work}, and algorithms to attribute authorship of revisioned content~\cite{flock2014wikiwho} could be used in future work to improve accurate identification of treatment editors.
  
  Furthermore, the non-standard nature of NPOV tags and variations in the tags limit the recall of detecting tag addition and tag removal revisions. To detect NPOV corrections, following the approach  of\citet{recasens2013linguistic}, we searched for occurrences of ``NPOV'', ``POV'', or any case variations (\cref{sec:identify-treatment-editors}) in the revision comments. However, other idiomatic terms such as ``point of view'', ``bias'', and ``weasel words'' could also be used to indicate NPOV corrections. Failing to detect these variants could result in an underestimation of the treatment effects. 
  
  \item \textbf{Controlling for platform-level changes over time.} While we limit the time range of our analyses to control for significant platform-level changes, there is a possibility for internal and external factors, such as influx of new editors, site upgrades, policy changes, to influence the findings. The article dataset used in this work was collected in 2013, and the policies and normative behavior in Wikipedia may have changed since then. Future work could consider replicating this work using datasets collected in different time periods to analyze any longitudinal changes in normative behavior in the platform. 

  \item \textbf{Assuming writing style changes in the short-term.}  A potential limitation with using ITS is that, for ITS, the outcomes are expected to change either relatively quickly after an intervention is implemented or after a clearly defined lag~\cite{bernal2017interrupted}. In our work, we assume that NPOV enforcement makes the articles and editors to improve in short-term. This could be a potential limitation, because in some cases, edits may occur long after the initial tagging.
  
  \item \textbf{Restricting article level analysis to unresolved articles.} Our article dataset was collected form the snapshot of articles in the \textit{NPOV dispute} in 2013. We use this single snapshot data and the articles in this dataset have an active NPOV tag (i.e., an unresolved NPOV issue). As a robustness check, we re-ran the regression for RQ1 using a subset of data after removing articles with long standing disputes and the effect of NPOV tagging treatment remains significant.
Using additional article data collection, the article level analysis could be expanded to cases where an NPOV tag was added and removed or where articles went through multiple NPOV tag additions/removals. This could provide insights about the effects after tag removal and repeated offenders, and future work could focus on this direction. From RQ1 we find that when an article has an active NPOV tag, subsequent revisions result in a decrease in bias language. Note that this finding does not imply that a tag is \emph{required} to reduce biased language in articles.
  \end{itemize}

\section{Conclusion}

While the Wikipedia NPOV tagging system has been in place for a long time, the impact of this tagging system has not been quantitatively measured. In this paper, we studied the effects of NPOV norm enforcement on Wikipedia using a corpus of NPOV-tagged articles and a set of style lexicons to characterize biased language. Focusing on the causal effect of NPOV norm enforcement both at the article and editor level, we found that NPOV tagging helps Wikipedia articles to converge to the accepted writing style, but no statistically significant improvements at editor level. 
Our findings suggest design improvements and interventions for Wikipedia and other online communities with strict language norms.

\section{Appendix A: Robustness Check for RQ1} \label{sec:appendix}
\begin{figure}[h]

\centering
\includegraphics[width=0.60\textwidth]{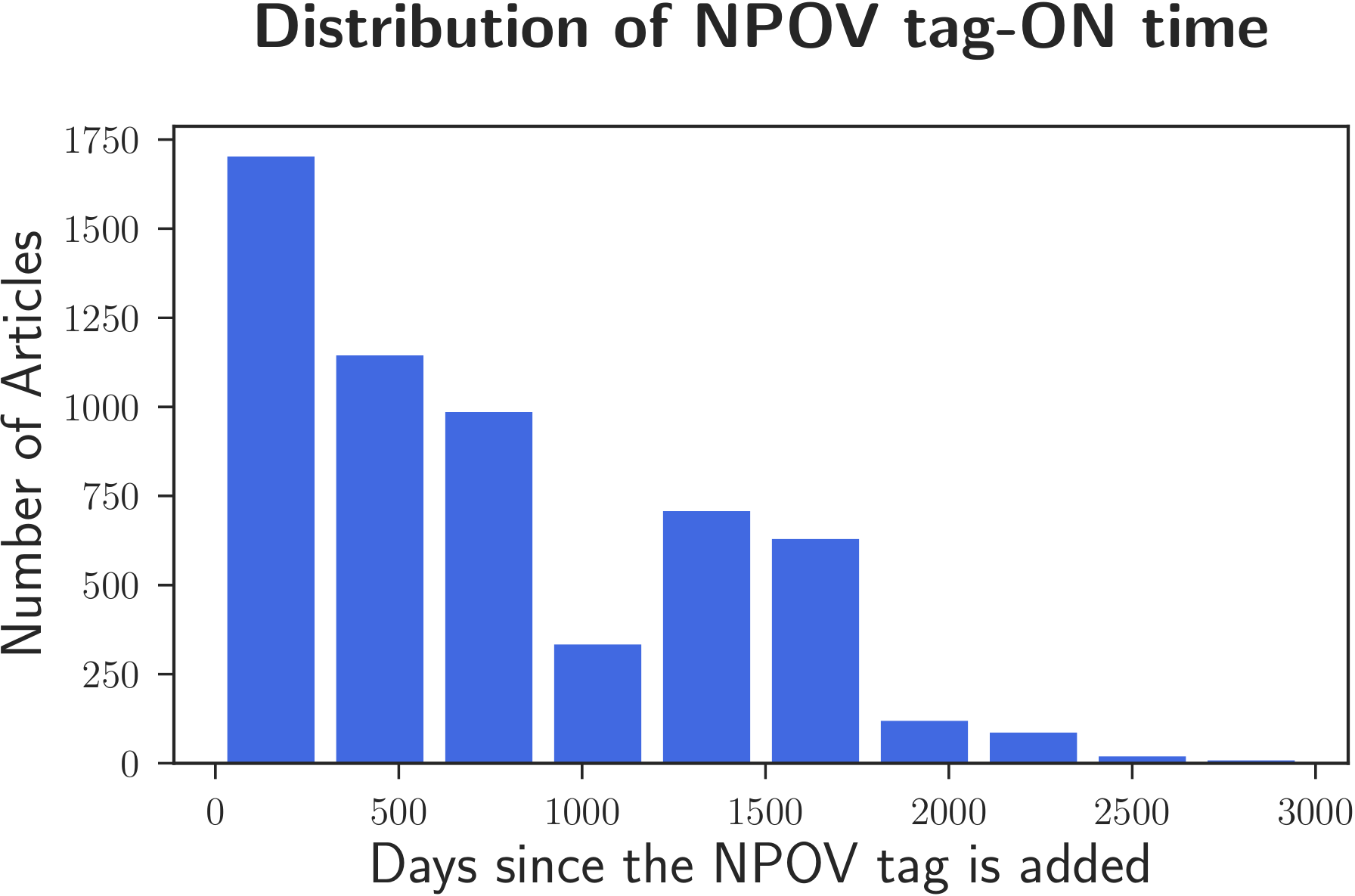}
\caption{The distribution of time-intervals between the addition of the NPOV tag and the data collection
time.}
\label{fig:article-tag-on-dist}
\end{figure}

The distribution of time-intervals between the addition of the NPOV tag and the data collection time is shown in \autoref{fig:article-tag-on-dist}. 

\autoref{tab:rq1-results-subset} shows results of the ITS analysis for RQ1 on the subset of data after removing articles which are NPOV tagged for more than three years.  The treatment effect remains significant after excluding these long-running disputes.


\begin{table}%
\caption{Results of RQ1 Analysis robustness check on the subset of data after removing articles which are tagged for more than three years. Article lexicon coverage is computed for the textual content of the article results from each revision. Coefficient $\beta_{3}$ indicates the change in slope after treatment. Statistical significance after correcting for multiple comparisons using \citet{benjamini1995controlling} adjustment for false discovery rate are shown; *: $p < 0.05$, **: $p < 0.01$, ***: $p < 0.001$. Percentage change is computed as the change in post-treatment lexicon coverage after 40 revisions.}
\label{tab:rq1-results-subset}
\begin{minipage}{\columnwidth}
\begin{center}
\begin{tabular}{lrrlr}
  \textbf{Lexicon} & \textbf{$\beta_{3}$}  & \textbf{$p$-value} &  & \textbf{\% change}\\
  \midrule
Words to Watch&-1.36e-05&3.79e-15&*** &-5.23\\
Hedges&-5.19e-06&3.66e-10&*** &-3.21\\
Assertives&-3.15e-06&4.39e-10&*** &-5.25\\
Positive Words&-2.62e-05&< 2e-16&*** &-4.13\\
Negative Words&-1.25e-05&4.07e-12&*** &-2.52\\
Factives&1.21e-06&3.16e-01& &-0.06\\
Implicatives&-9.84e-07&3.66e-02&* &-2.08\\
Report Verbs&-5.92e-06&1.26e-04&** &-3.79\\
Strong Subjectives&-3.75e-05&< 2e-16&*** &-8.74\\
Weak Subjectives& -2.80e-05&< 2e-16&*** &-1.55\\

\end{tabular}

\end{center}
\bigskip\centering
\end{minipage}

\end{table}%

\newpage

\section{Acknowledgments}
We thank the anonymous reviewers for their helpful and constructive feedback. We thank Amy Bruckman for helpful discussions; Scott Appling, Stevie Chancellor, Sandeep Soni, and Ian Stewart for their valuable feedback on the paper. This research was supported by Air Force Office of Scientific Research award FA9550-14-1-0379,  by National Institutes of Health award R01-GM112697, and by National Science Foundation awards 1452443 and 1111142.









\bibliographystyle{ACM-Reference-Format}
\bibliography{npov}

\end{document}